\documentclass[a4paper]{article}

\usepackage[english]{babel}
\usepackage[utf8]{inputenc}
\usepackage{amsmath}
\usepackage{graphicx}
\usepackage[colorinlistoftodos]{todonotes}
\usepackage{amssymb,amsthm}
\usepackage{hyperref}
\usepackage{verbatim}
\usepackage{multirow}
\usepackage{caption}
\usepackage{hyperref}

\DeclareMathOperator*{\argmax}{arg\,max}

\title{Deep Learning Models for Automatic Summarization\\
\begin{large}
The Next Big Thing in NLP?
\end{large}
}

\author{
Pirmin Lemberger\\
{\small \texttt{p.lemberger@groupeonepoint.com}}\\
\\
   {\small \textit{onepoint}}\\
   {\small 29 rue des Sablons, 75116 Paris} \\
   {\small \texttt{groupeonepoint.com}}\\ \\
   }
\date{\today}

\begin{document}
\maketitle

\newcommand{\bx}{\mathbf{x}}
\newcommand{\by}{\mathbf{y}}
\newcommand{\bz}{\mathbf{z}}
\newcommand{\xfeatures}{x_1 \dots, x_p}
\newcommand{\Xset}{\mathcal{X}}
\newcommand{\Yset}{\mathcal{Y}}
\newcommand{\R}{\mathbb{R}}
\newcommand{\E}{\mathbb{E}}


\begin{abstract}
Text summarization is an NLP task which aims to convert a textual document into a shorter one while keeping as much meaning as possible. This pedagogical article reviews a number of recent Deep Learning architectures that have helped to advance research in this field. We will discuss in particular applications of pointer networks, hierarchical Transformers and Reinforcement Learning. We assume basic knowledge of Seq2Seq architecture and Transformer networks within NLP.
\\ 
\end{abstract}

\section{Perhaps the most helpful NLP task of all}
For over a quarter of century we have been able to search the web by querying a search engine using a couple of relevant keywords. Without such a tool the internet would be nothing but useless garbage dump of data. In 1998 Google’s \href{https://en.wikipedia.org/wiki/PageRank}{{PageRank algorithm}} redefined what we can expect as far as relevance of search results is concerned. More recently some \href{https://searchengineland.com/welcome-bert-google-artificial-intelligence-for-understanding-search-queries-323976}{{semantic processing}} has been added to the wizardry that helps the engine to interpret a query that was expressed in plain language. In a not too distant future we may perhaps pinpoint documents by engaging a short Q\&A kind of conversation with a search engine, just as we would with a bookseller. There is an important difference though between a bookseller and a search engine. If you are hesitant about which book you should read you could try to ask the bookseller to summarize it for you in few of sentences. 
\\

This kind summarization task has long seemed totally out of reach within the classic rule-based NLP approaches and neither was it not considered realistic in foreseeable future. But, slowly, things are now changing with recent progress in Deep Learning models for NLP. For the moment just imagine you had a drop down list next to the input field of your favorite search engine that would allow you to set the length of an automatic summary for a given document. Say, a 1 sentence, a 10 sentences or a one page summary. Would that be helpful? Actually it is quite possible that it could quickly prove so useful that it could become ubiquitous. Besides improving document search it could also help in a multitude of other tasks. For instance it could help scientists keep up with a dizzying flow of publications in fields like medicine or AI. More prosaically it could help producing \href{https://arxiv.org/abs/1807.08000}{{short product descriptions} for online stores} with catalogues too large to be handled manually. Many more examples of applications of automatic summarization are described for instance \href{https://blog.frase.io/20-applications-of-automatic-summarization-in-the-enterprise/}{{here}}.
\\

For larger documents with several hundreds of pages like novels such generic summarization tools still belong to the realm of science fiction. However, thanks to the ever surprising flexibility of Deep Learning models, the wait may not be that long for tools that could summarize one- or two-page documents in a few sentences, at least within specific areas of knowledge. The aim of this article is to describe recent data sets \cite{BIGPATENT, TalkSumm} and Deep Learning architectures \cite{GetToThePoint, HIBERT, DeepRLModel} that have brought us a little closer to the goal.

\section{A difficult task}
The summarizing task is difficult for a number of reasons, some of which are common to other NLP tasks like translation for instance:
\begin{itemize}
\item For a given document \textbf{there is no summary which is objectively the best}. As a general rule, many of them that would be judged equally good by a human.
\item It is hard to define precisely \textbf{what a good summary is} and what score we should use for its evaluation.
\item Good training data has long been \textbf{scarce} and \textbf{expensive} to collect.
\end{itemize}
Human evaluation of a summary is subjective and involves judgments like style, coherence, completeness and readability. Unfortunately no score is currently known which is both easy to compute and faithful to human judgment. The \href{https://rxnlp.com/how-rouge-works-for-evaluation-of-summarization-tasks/#.Xqbe75MzbUJ}{{ROUGE score}} \cite{ROUGE} is the best we have but it has obvious shortcomings as we shall see. ROUGE simply counts the number of words, or $n$-grams, that are common to the summary produced by a machine and a reference summary written by a human. More precisely it reports a combination of the corresponding \textbf{recall}:
\[
	\mathrm{recall}:=\frac{\mathrm{\# overlapping} \;n–\mathrm{grams}}{\mathrm{\# words\;in\;the\;reference\;summary}},
\]
and \textbf{precision}:
\[
	\mathrm{precision}:=\frac{\mathrm{\# overlapping} \;n–\mathrm{grams}}{\mathrm{\# words\;in\;the\;machine\;summary}}.
\]

The combination reported in ROUGE-$n$ is their geometric mean (known as the F1 score). Although the ROUGE score does not faithfully reflect a human judgment it has the advantage of computational simplicity and it takes into account some of the flexibility associated with the multiple summaries that could result by rearranging words within a valid summary.
\\

There are two types of summarization systems: 
\begin{itemize}
	\item \textbf{Extractive summarization} systems select a number of segments from the source document to make up a summary. The advantage of this approach is that the resulting summary is guaranteed to be grammatically correct. In general extractive systems achiever high ROUGE scores and are more reliable than the option we discuss next.
	\item \textbf{Abstractive summarization systems} on the other hand generate their own words and sentences to reformulate the meaning of the source as a human writer would do. They can be viewed as compression systems that attempt to preserve meaning. This latter kind of systems is obviously more difficult to develop because it involves the ability to paraphrase information and to include external knowledge. 
\end{itemize}
We will describe instances of both kinds below.

\section{More and better Data}
Until recently the main data set used for training summarization models was the \href{https://github.com/abisee/cnn-dailymail}{\textbf{{CNN / Daily Mail data set}}} which contains 300,000 examples of news article paired with their multiline summary. A detailed examination \cite{BIGPATENT}, however, has revealed various limitations in this data set that could bias the evaluation of the ability of a system to perform text summarization. It turned out for instance that useful information is spread unevenly across the source, namely mostly at the beginning of the documents. Moreover, many summaries contain large fragments of the source. This is certainly not the best way for teaching a system how to produce good abstractive summaries.
\\
 
But things have changed recently. The \href{https://arxiv.org/abs/1906.03741}{ {\textbf{BigPatent dataset}}} \cite{BIGPATENT} for instance contains 1.3 millions patent documents together with their summaries that alleviate most of the above shortcomings.
\\

A novel approach to produce ever growing data sets for training summarization models uses video transcripts of talks given at international scientific conferences. The basic assumption here is that these transcripts make a good starting point for producing high quality summaries of scientific papers. The transcript itself is not directly the summary of a paper. Rather, the authors of the \href{https://arxiv.org/abs/1906.01351}{{\textbf{TalkSumm method}}} \cite{TalkSumm} propose to create a summary by retrieving a sequence of relevant sentences from the paper presented in the talk. A sentence is deemed relevant depending on how many words the speaker uses to describe it in her talk, assuming that she has a given sentence of the paper in mind at any given point in time.

\section{Clever architectures, improved cost functions}
In this section we describe 3 neural network models that have be developed recently for the summarization task. The aim here is not completeness of course but merely to illustrate the diversity of ideas which have been proposed to tackle this fundamental NLP problem.
\\

The basic neural network architectures that make it possible to learn this kind of task are the \href{https://guillaumegenthial.github.io/sequence-to-sequence.html}{{\textbf{Seq2Seq architectures}}}, the \href{https://colah.github.io/posts/2015-08-Understanding-LSTMs/}{{\textbf{LSTM}}} recurrent neural networks (RNN), the \href{http://jalammar.github.io/illustrated-bert/}{{\textbf{BERT}}} and the \href{http://jalammar.github.io/illustrated-transformer/}{{\textbf{Transformer}}} models as well as the \href{https://arxiv.org/abs/1409.0473}{{\textbf{attention mechanism}}}.
\\

\begin{figure}[h]
\centering
\includegraphics[scale=0.9]{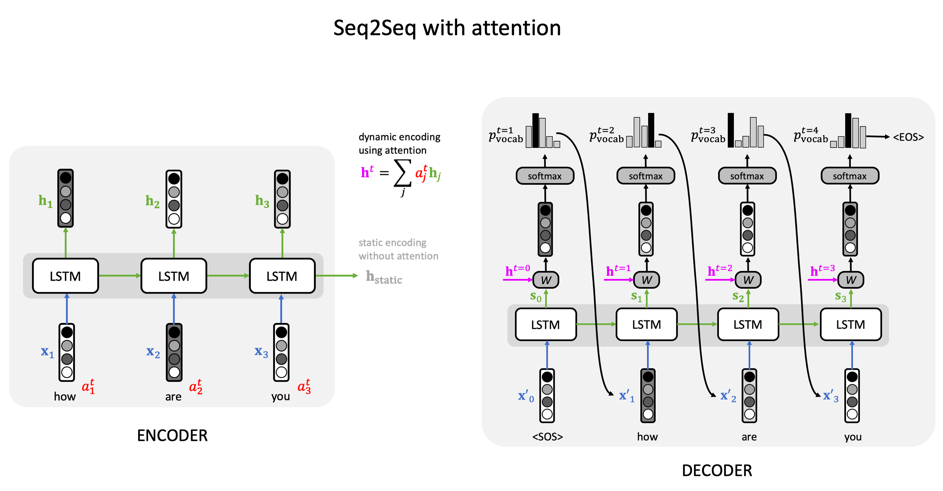}
\caption{{\small Basic Seq2Seq encoder-decoder architecture with attention. The $\mathbf{x}_i$} are the input token embeddings, the coeffici $a_i^t$ are the attention weights at step $t$, the $\mathbf{h}_i$ are the context vectors, $\mathbf{h}^t$ is the sentence embedding at step $t$ obtained by weighting the context vectors with the attention weights, $\mathbf{s}_i$ are the decoder states, $\mathbf{x}'_i$ are the embeddings of the generated token (at inference time) or ground truth tokens (at training time when using teacher forcing). At last $p^t_\mathrm{vocab}$ is the probability distribution at time $t$ over a fixed vocabulary.}
\label{seq2seq_fig}
\end{figure}

For the readers unfamiliar with any of these topics we recommend the above links which will provide excellent introductions to each of them. Figure \ref{seq2seq_fig} represents the Seq2Seq architecture which converts a sequence of tokens into another sequence with a possibly different length. It defines the vectors we will refer to when talking about Seq2Seq.
\\

\begin{figure}[h]
\centering
\includegraphics[scale=0.8]{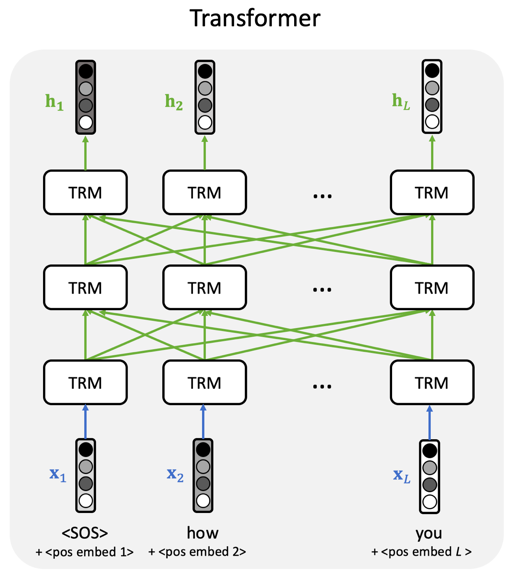}
\caption{{\small BERT as the encoder part of the Transformer architecture. The core idea behind the Transformer is a smart implementation of attention mechanism that allows computations to be parallelized efficiently on GPU’s, something that was not possible with classical RNN. Each input vector $\mathbf{x}_j$ is a sum of a token embedding and a  position embedding. The outputs $\mathbf{h}_i$ are context aware token embeddings.}}
\label{Transformer_fig}
\end{figure}

Figure \ref{Transformer_fig} sketches a Transformer network with the self-attention dependencies between embeddings an hidden vectors. Roughly speaking a transformer converts a sequence of token embeddings $\mathbf{x}_i$ into another sequence of context aware embeddings $\mathbf{h}_i$. The input vectors $\mathbf{x}_i$ also typically include positional information. This is needed in contrast to RNN networks because of the permutation symmetry of inputs in a Transformer.

\subsection{Summarizing without stuttering}
\begin{figure}[h]
\centering
\includegraphics[scale=0.8]{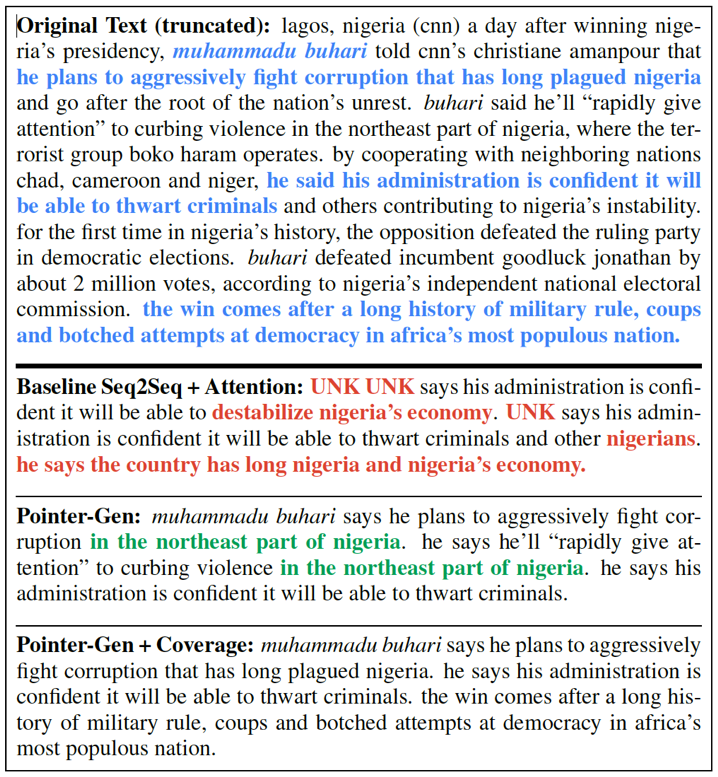}
\caption{{\small The last section ‘‘Pointer-Gen + Coverage'' contains the output of the system proposed in \cite{GetToThePoint}. The fragments used in the summary are shown in blue, factual errors in red and unnecessary repetitions in green.}}
\label{summary_example_fig}
\end{figure}

The first architecture we present addresses the abstractive summarization task \cite{GetToThePoint}. Early attempts to apply vanilla Seq2Seq architectures to the summarization revealed a number of issues with this straightforward approach:
\begin{itemize}
	\item \textbf{Factual details} in the source document, like dates, locations or phone numbers, were often reproduced incorrectly in the summary.
	\item A \textbf{finite vocabulary} prevents some words like proper names from being taken into accounts.
    \item Unnecessary repetitions of source fragments often occur, in other words \textbf{the model tends to stutter}.
\end{itemize}
Figure \ref{summary_example_fig} shows examples of these unwanted behaviors.
The authors in \cite{GetToThePoint} propose two improvements over the vanilla Seq2Seq with attention mechanism to mitigate these shortcomings.
\\

First, to overcome the finite vocabulary limitation they allow the network to copy a word directly from the source and use it in the summary when needed. The precise mechanism to do this is called a \textbf{pointer network}. Remember that in a vanilla Seq2Seq network the decoder computes a probability distribution $p^t_\mathrm{vocab}(w)$, at each time step $t$, over the words $w$ in a fixed finite vocabulary. As usual $p^t_\mathrm{vocab}$ is computed with a softmax layer that takes the attention context vector $\mathbf{h}^t$ and the decoder state $\mathbf{s}_t$ as inputs. In a pointer network an additional \textbf{copy probability} $p_\mathrm{copy}$ is computed which represents the probability that a word should be copied from the source rather than generated by the decoder. The probability $p_\mathrm{copy}$ is computed using a sigmoid layer having $\mathbf{h}^t$, $\mathbf{s}_t$ and $\mathbf{x}_t$ vectors as inputs (see figure \ref{seq2seq_fig}). Which word should actually be copied is determined by the attention weights $a_i^t$ that the decoder puts at time $t$ to each word $w_i$ in the source. Putting it all together, the full probability for the model to produce the word $w$ is thus given by the following mixture:

\[
	p^t(w):=\left(1-p_{\mathrm{copy}} \right)
	p^t_\mathrm{vocab}(w) + 
	p_\mathrm{copy}\sum_{i:w_i=w}a^t_i.
\]

Second, to avoid repeating the same segments the authors define a \textbf{coverage vector} $\mathbf{c}^t$ at each time step $t$ which estimates the amount of attention that each word $w_i$ in the source has received from the decoder until time $t$:
\[
	c_i^t := \sum_{s=0}^{t-1}a_i^s.
\]
This coverage vector is then used in two different places within the network. First it is used to inform the attention mechanism in charge of computing the attention weights $a^t_i$ (in addition to the usual dependence on the encoder context vector $\mathbf{h}_i$ for the word $w_i$ and the decoder state $\mathbf{s}_t$). The decoder is thus aware of the words it has already been paying attention to. Second it is used to correct the loss function. Remember that at time step $t$ the weight $a^t_i$ is the attention put on word $w_i$ while $c^t_i$ is the attention this same word has received in the past. If the word $w_i$ receives more attention at time $t$ than it has already received in the past, that is if $a^t_i > c^t_i$, then the cost function should penalize large values of $c^t_i$ and also the other way around. To penalize attention to repeated words one defines an additional term in the \textbf{loss function} at time step $t$ as a sum over input tokens:
\[
	L^t_{\mathrm{coverage}}:=\sum_i\min\left(a_i^t, c_i^t \right).
\]
This is then added (with an additional hyperparameter $\lambda$) to the usual negative log likelihood $L^t_{\mathrm{ML}}:=-\log p^t(w_t^*)$ of the target word $w^*_t$ in the train set: 
\[
	L^t := L^t_{\mathrm{ML}} + \lambda L^t_\mathrm{coverage}.
\]
Results with and without these additional tricks are shown in figure \ref{summary_example_fig}.

\subsection{Documents as sequences of contextualized sentences} 
Our next example illustrates recent ideas that defined a new SOTA for the extractive summary task. It builds directly upon a key idea that lead to the 
\href{http://jalammar.github.io/illustrated-bert/}{{\textbf{BERT model}}}
 in 2018, namely that of transfer learning based on a clever pretraining task for a \href{http://jalammar.github.io/illustrated-transformer/}{{\textbf{Transformer}}} encoder. Let’s go into a little more detail and summarize the \href{https://arxiv.org/abs/1905.06566}{{\textbf{HIBERT}}} architecture for document summarization \cite{HIBERT} .
\\

\begin{figure}[h]
\centering
\includegraphics[scale=0.22]{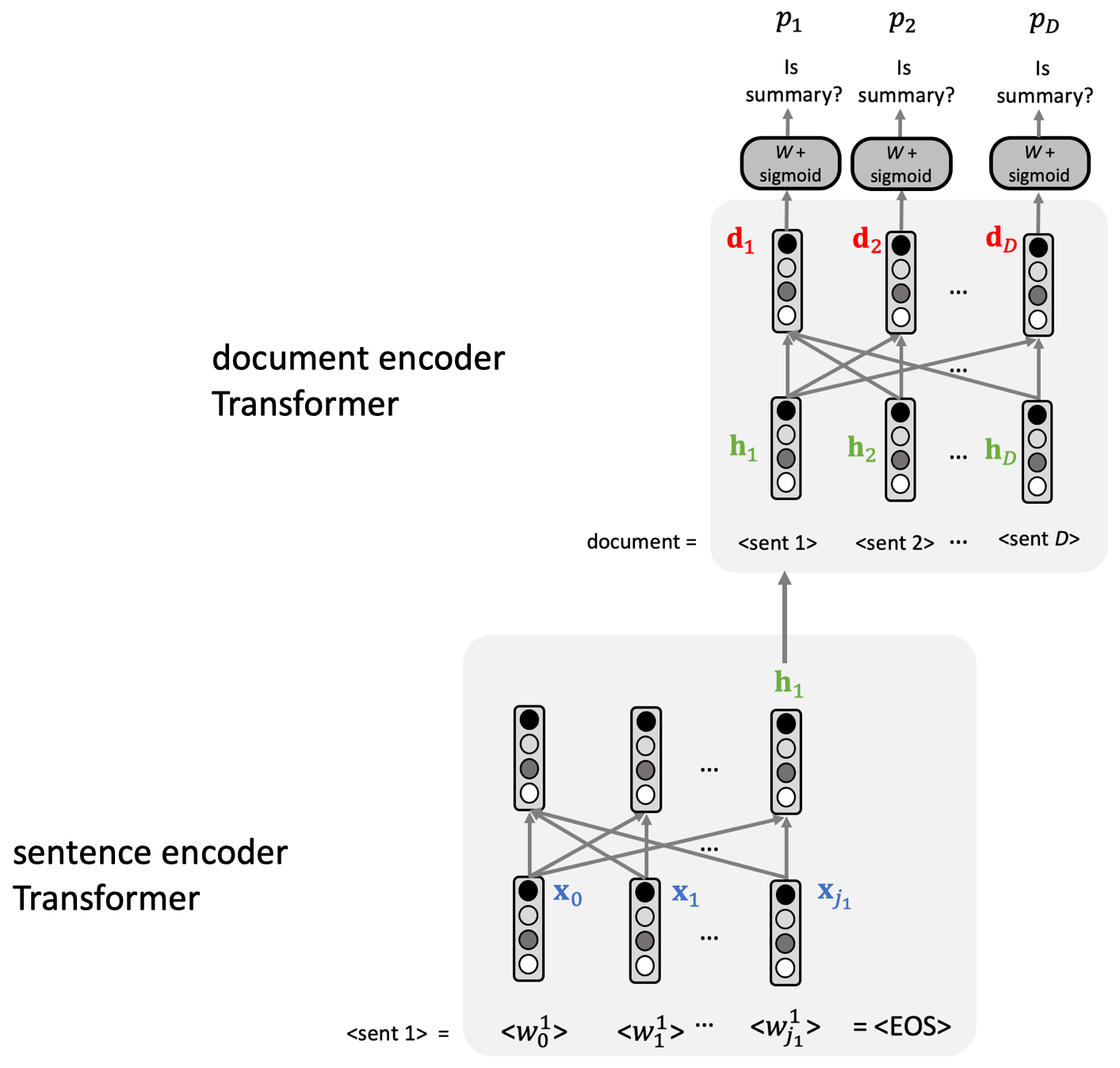}
\caption{{\small The HIBERT architecture involves a hierarchy of two Transformer encoders used to classify each sentence within a document as being part of the summary or not.}}
\label{HIBERT_fig}
\end{figure}

The basic observation is that extractive classification can be cast as a \textbf{sentence tagging problem}: simply train a model to identify which sentence in a document should be kept to make up summary! For this purpose the HIBERT architecture uses two nested encoder Transformers as illustrated in figure \ref{HIBERT_fig}.
\\

The first Transformer encoder at the bottom is a classic \textbf{sentence encoder} that transforms the sequence of words $(w_0^k, w_1^k,…, w_{j_k}^k)$ that make up the $k^\mathrm{th}$ sentence of the document to be summarized into a sentence embedding $\mathbf{h}_k$. This vector is conventionally identified as the context vector above the end of sentence token $\langle\mathrm{EOS}\rangle$. 
\\

The second Transformer encoder which sits on the top is a \textbf{document encoder} that transforms the sequence of sentence embeddings $(\mathbf{h}_1, \mathbf{h}_2,…, \mathbf{h}_D)$ into a sequence of \textbf{document aware sentence embeddings} $(\mathbf{d}_1, \mathbf{d}_2,…, \mathbf{d}_D)$. These embeddings are in turn converted into a sequence of probabilities $(p_1, p_2,…, p_D)$ where $p_j$ is the probability that the $j^\mathrm{th}$ sentence should be part of the summary.
\\

Training such a complex hierarchical network from scratch is impractical because it would require an unrealistic amount of document-summary pairs. As is well known, the best strategy to train such a complex network with a limited amount of data is to use \textbf{transfer learning}. For this purpose the HIBERT architecture is first pretrained on an auxiliary task which consists in predicting sentences that are randomly masked (15\% of them) within in a large corpus of documents:

\begin{quote}
\textit{Turing was an English mathematician. \textcolor{cyan}{He was highly influential}. Turing is widely considered to be the father of artificial intelligence.}
\end{quote}

\begin{figure}[h]
\centering
\includegraphics[scale=0.18]{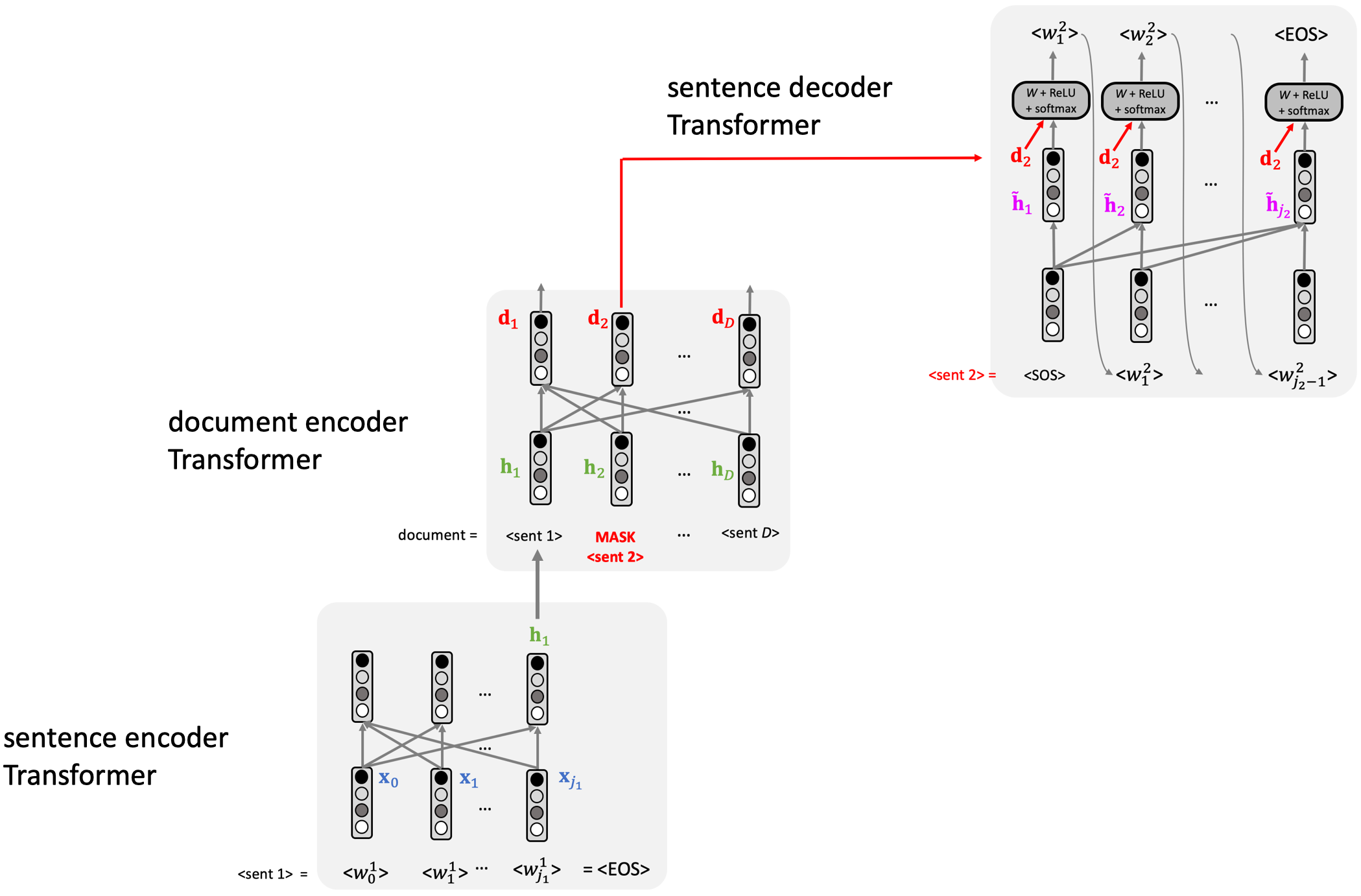}
\caption{{\small The architecture used for the masked sentence prediction task. A sentence Transformer decoder is added on top of the HIBERT architecture to recover the words of a masked sentence using the information encapsulated in its document aware embedding $\mathbf{d}_k$.}}
\label{HIBERT_pretraining_fig}
\end{figure}

\begin{quote}
\textit{Turing was an English mathematician. \textcolor{red}{[MASK] [MASK] [MASK] [MASK]}. Turing is widely considered to be the father of artificial intelligence.}
\end{quote}
Figure \ref{HIBERT_pretraining_fig} shows the architecture used for this \textbf{masked sentence prediction task}. It adds a Transformer decoder on top of the HIBERT architecture in order to convert the document aware sentence embedding $\mathbf{d}_k$ into the sequence of words $(w_0^k, w_1^k,…, w_{j-1}^k)$ of the $k^\mathrm{th}$ sentence which was masked. To generate the word at step $i$ the decoder uses both its context vector $\tilde{\mathbf{h}}_i$ and the document aware sentence embedding $\mathbf{d}_k$ from the document encoder.
\\

Trained this way the network gathers a large amount of semantic knowledge without requiring any expansive labeling procedure. In a second stage, leveraging what it learned during the pretraining task, the network is fine-tuned on the actual target task, namely summarization as a sentence binary tagging task as in figure 4 describes.  
\\

This masked sentence prediction task is obviously reminiscent, on a sentence level, of the \textbf{masked language model} (MLM) used for pretraining the original BERT model. Remember that the MLM task consisted in recovering randomly masked words within sentences. 

\subsection{Reinforcement Learning comes to the rescue}
As we explained earlier one central issue with the summarization task is the lack of a unique best summary. The ROUGE score takes this into account, up to some level, because it ignores the order of the words (or $n$-grams) within the generated summary. Therefore the cost function we would actually like to minimize should be something like this ROUGE score, or at least the final loss function should include such a term. This is the strategy that was followed in the last work \cite{DeepRLModel} we present here which, again, concerns abstractive summarization.
\\

The problem with a score like ROUGE is that for any sequence of words $(w_1,…, w_j)$ generated by the decoder, it is constant with respect to the parameters $\mathbf{\theta} $ of the network, thus making backpropagation impossible. The situation is not hopeless though because the expectation of the ROUGE score for sentences $(w_1,…, w_j)$, sampled from the joint probability distribution $p_\theta(w_1,…, w_j)$ defined by the generator \textit{is} indeed a \textbf{differentiable} function of those parameters $\theta$! The way to go is clear then. Just minimize the loss $L_\mathrm{RL}(\theta)$ defined by that expectation:
\[
	L_\mathrm{RL}(\theta):=-\mathbb{E}_{\left(w_1,...,w_j \right)\sim p_\theta}	
	\left[\mathrm{ROUGE}\left(w_1,...,w_j \right) \right]
\]
Actually we can view the generator of a Seq2Seq model as a \textbf{Reinforcement Learning} (RL) agent whose action at time step $t$ is to generates a word $w_t$ depending on an inner state $\mathbf{s}_t$ which encapsulates the history from previous actions. From here on we just need to open a book on RL \cite{SuttonAndBartoRL} to learn how to minimize $L_\mathrm{RL}(\theta)$. A basic result in RL, known as the \textbf{Policy Gradient Theorem}, states that the gradient of $L_\mathrm{RL}(\theta)$:
\[
	\nabla_\theta L_\mathrm{RL}:=-\mathbb{E}_{\left(w_1,...,w_j \right)\sim p_\theta}	
	\left[\mathrm{ROUGE}\left(w_1,...,w_j \right) \nabla_\theta\log
	p_\theta\left(w_1,...,w_j\right)\right]
\]
where
\[
	\log
	p_\theta\left(w_1,...,w_j\right):=\sum_{t=1}^j
	\log p_\theta\left(w_t|w_1,...,w_{t-1}\right)
\]
and the last index $j$ is that of the $\langle \mathrm{EOS}\rangle$ token. The \textbf{REINFORCE} algorithm approximates the above expectation with a single sample $(w_1,…, w_j)$ from the distribution $p_\theta(w_1,…, w_j)$ computed by the generator:
\[
	\nabla_\theta L_\mathrm{RL}\approx
	-\left[ \mathrm{ROUGE}(w_1,...,w_j) \nabla_\theta 
	\sum_{t=1}^j \log p_\theta(w_t|w_1,...,w_{t-1}) \right]
\]
In practice scores like ROUGE can have a \textbf{large variance} which hinders convergence of the gradient descent. Fortunately, we can enhance the speed of convergence by comparing ROUGE$(w_1,…, w_j)$ to a \textbf{baseline} $b$ which is independent of $(w_1,…, w_j)$. This does not change the gradient of $L_\mathrm{RL}$ as can readily be verified but it can considerably reduce the variance \cite{SuttonAndBartoRL} and thus dramatically improve convergence:
\[
	\nabla_\theta L_\mathrm{RL}\approx
	-\left[\left[\mathrm{ROUGE}(w_1,...,w_j)-b\right] \nabla_\theta 
	\sum_{t=1}^j \log p_\theta(w_t|w_1,...,w_{t-1})\right]
\]
The main point thus is to find an appropriate baseline. The idea in the work we are discussing \cite{DeepRLModel} is to take the baseline $b$ equal to the ROUGE score of the sequence of words $(\hat{w}_1,...,\hat{w}_j)$ the generator actually generates at inference time. Remember that this is the sequence of words that successively maximize the conditional probabilities as computed by the softmax of the decoder at each step $t$:
\[
	\hat{w}_t:=\argmax_{w_t}p_\theta(w_t|\hat{w}_1,...,\hat{w}_{t-1}).
\]
This choice for the baseline $b$ is called \textbf{self-critical sequence training} (SCST). Altogether the reinforcement loss term thus reads:
\[
	L_\mathrm{RL}\approx
	\left[\left[\mathrm{ROUGE}(\hat{w}_1,...,\hat{w}_j)-\mathrm{ROUGE}		
	(w_1,...,w_j)\right] 
	\sum_{t=1}^j \log p_\theta(w_t|w_1,...,w_{t-1})\right].
\]
where
\[
\begin{split}
   & w_t \mathrm{\; is\;  sampled\; successively\; from\;} p_\theta(w_t|w_1,...,w_{t-1})\mathrm{\;for\;}t=1,...,j, \\
   & \hat{w}_t \mathrm{\;successively\;maximizes\;} p_\theta(w_t|\hat{w}_1,...,\hat{w}_{t-1})\mathrm{\;for\;}t=1,...,j.
\end{split}
\]
This loss term as we can see prompts $p_\theta$ to generate word sequences $(w_1,...,w_j)$ whose ROUGE score is larger than that of the sequence $(\hat{w}_1,...,\hat{w}_j)$ that was currently generated by the decoder. 
\\

There are two benefits for including such a SCST reinforcement learning term $L_\mathrm{RL}$ in the loss function. The first, which motivated the construction $L_\mathrm{RL}$ of in the first place, is that it makes it possible to use a non-differentiable score like ROUGE within a stochastic gradient descent training procedure. The second benefit is that it also cures the so called \textbf{exposure bias}. Exposure bias results from the classic \textbf{teacher forcing} procedure that is typically used to train a Seq2Seq model. This procedure trains the decoder RNN using the ground truth words $(w^*_1,...,w^*_j)$ from the train set while at inference time the decoder must of course use its own generated tokens $(\hat{w}_1,...,\hat{w}_j)$, which could therefore result in an accumulation of errors. The SCST choice for the baseline $b$ amounts to train the decoder using the distribution it will actually see at inference time.
\\

The final loss function used is a weighted sum of the \textbf{reinforcement learning loss} $L_\mathrm{RL}$ and a standard \textbf{maximum likelihood objective} $L_\mathrm{ML}$. The former takes into account the non-uniqueness of summaries, at least up to some point, but by itself it is certainly not an incentive for the model to produce readable messages. The latter on the other hand favors readable sentences as it is  basically defines a language model.
\\

In order to avoid repetitions, the authors also use an enhanced attention mechanism that involves a pointer network similar to the one we described in the first example \cite{GetToThePoint}.

\section{What’s next?}
The three models we described in the previous section all use Deep Learning and therefore implement a purely statistical approach to the summarization task. Recent research also tries to find \textbf{better loss functions}. Researchers at \href{https://recital.ai/}{{ReciTAL}} for instance explore the interesting idea that a good summary should answer questions as well as the original text allows \cite{ReciTAL}. On the whole, these models work indeed surprisingly well for short documents. But can we reasonably expect to build a system that could summarize a 300 pages novel in a page using techniques that only rely on crunching huge amounts of textual data? This is far from obvious. Abstract summarization should in principle be able to leverage real world knowledge to make sense of a document or a book to be summarized. It is unlikely though that language models alone, even when initialized with clever pretraining tasks, can ever capture such common sense which is more likely collected by \textbf{sensory experience}. One short term possibility for building useful summarization tools could be to narrow their scope down to specific areas of expertise where knowledge basis or ontologies are already available. A more radical step towards building system with better “real world understanding” could arise from \textbf{multimodal learners} designed to aggregate audio, video and text modalities, from movies from instance. Promising results have already been obtained along this path \cite{Multimodal}.

\section*{Acknowledgments}
I would like here to thank Thomas Scialom, researcher at \href{https://recital.ai/}{ReciTAL}, who kindly share his knowledge with me by pointing my attention to his \href{https://github.com/recitalAI/summarizing_summarization}{Summarizing Summarization} page \cite{SumSumTS}. This helped me kick start my exploration of Deep Learning summarization models.

\end{document}